\documentclass[11pt,a4paper]{article}
\usepackage[hyperref]{acl2020}
\usepackage{times}
\usepackage{latexsym}

\usepackage{url}

\usepackage{array}
\usepackage{color}

\usepackage{graphicx}
\usepackage[font=normalsize,position=bottom]{subfig}
\usepackage[noend]{algpseudocode}
\usepackage{multirow}
\usepackage{amssymb}
\usepackage{bm}
\usepackage{balance}
\usepackage{arydshln}
\usepackage{mathtools}
\usepackage{soul}
\usepackage[ruled]{algorithm2e}

\usepackage{enumitem}

\usepackage{booktabs}
\usepackage{amsmath,amsfonts}
\usepackage{xspace}

\usepackage{diagbox}

\usepackage{bbm}

\usepackage{xspace}
\usepackage{textcomp}

\usepackage{tabularx}
\usepackage{CJKutf8}
\usepackage[toc,page]{appendix}
\usepackage{verbatim}

\def\argmax{\mathop{\rm argmax}}%
\def\argmin{\mathop{\rm argmin}}%

\newcommand{\eat}[1]{} 

\def\S{\mathcal{S}} %
\def\cS{\vert \mathcal{S} \vert} %

\def\bx{\mathbf{x}} %
\def\bxp{\mathbf{x}^{\prime}} %
\def\bxpp{\mathbf{x}^{\prime\prime}} %
\def\tbx{\tilde{\mathbf{x}}} %
\def\hbx{\hat{\mathbf{x}}} %
\def\ex{e(\mathbf{x})}
\def\tex{e({\tilde{\mathbf{x}}})}

\def\by{\mathbf{y}} %
\def\byp{\mathbf{y}^{\prime}} %
\def\bypp{\mathbf{y}^{\prime\prime}} %
\def\tby{\tilde{\mathbf{y}}} %
\def\hby{\hat{\mathbf{y}}} %
\def\ey{e(\mathbf{y})}
\def\tey{e({\tilde{\mathbf{y}}})}

\def\Axy{A_{(\mathbf{x}, \mathbf{y})}} %

\newcommand{\eg}{\emph{e.g. }} 
\newcommand{\ie}{\emph{i.e. }}

\newcommand{\cn}[1]{\begin{CJK*}{UTF8}{gbsn}{#1}\end{CJK*}}

\newcommand{\fancyname}{AdvAug}
\newcommand{\mixup}{\textit{mixup}\xspace} 

\usepackage{microtype}

\aclfinalcopy 
\def\aclpaperid{2508} 

\title{\fancyname: Robust Adversarial Augmentation for \\ Neural Machine Translation
}
\author{Yong Cheng, Lu Jiang, Wolfgang Macherey and Jacob Eisenstein \\
Google Research \\
\texttt{\{chengyong, lujiang, wmach, jeisenstein\}@google.com}\\}

\date{}

\begin{document}
\maketitle
\begin{abstract}
In this paper, we propose a new adversarial augmentation method for Neural Machine Translation (NMT).
The main idea is to minimize the vicinal risk over virtual sentences sampled from two vicinity distributions, of which the crucial one is a novel vicinity distribution for adversarial sentences that describes a smooth interpolated embedding space centered around observed training sentence pairs. We then discuss our approach, {\em \fancyname}, to train NMT models using the embeddings of virtual sentences in sequence-to-sequence learning. Experiments on Chinese-English, English-French, and English-German translation benchmarks show that {\em \fancyname} achieves significant improvements over the Transformer (up to 4.9 BLEU points), and substantially outperforms other data augmentation techniques (\eg back-translation) without using extra corpora.
\end{abstract}

\section{Introduction}
Recent work in neural machine translation \cite{Bahdanau:15,Gehring:17,Vaswani:17} has led to dramatic improvements in both research and commercial systems~\cite{Wu:16}.
However, a key weakness of contemporary systems is that performance can drop dramatically when they are exposed to input perturbations \cite{Belinkov:17, Cheng:19}, even when these perturbations are not strong enough to alter the meaning of the input sentence. 
Consider a Chinese sentence, 
``zhejia feiji meiyou zhuangshang zhujia \textbf{huo} yiyuan, shizai shi qiji''. If we change the word ``\textbf{huo}~(\cn{或})'' to its synonym``\textbf{ji}~(\cn{及})'', the Transformer model will generate contradictory results of ``It was indeed a miracle that the plane \textbf{did not touch down} at home or hospital." versus ``It was a miracle that the plane \textbf{landed} at home and hospital.''
Such perturbations can readily be found in many public benchmarks and real-world applications. 
This lack of stability not only lowers translation quality but also inhibits applications in more sensitive scenarios.

At the root of this problem are two interrelated issues: first, machine translation training sets are insufficiently diverse, and second, NMT architectures are powerful enough to overfit --- and, in extreme cases, memorize --- the observed training examples, without learning to generalize to unseen perturbed examples.
One potential solution is data augmentation which introduces noise to make the NMT model training more robust.
In general, two types of noise can be distinguished: (1) continuous noise which is modeled as a real-valued vector applied to word embeddings~\cite{Miyato:15,Miyato:17,Cheng:18,Sano:19}, and (2) discrete noise which adds, deletes, and/or replaces characters or words in the observed sentences~\cite{Belinkov:17, Sperber:17, Ebrahimi:18b, Michel:19, Cheng:19, Karpukhin:19}. In both cases, the challenge is to ensure that the noisy examples are still semantically valid translation pairs.
In the case of continuous noise, it only ensures that the noise vector lies within an $L_2$-norm ball but does not guarantee to maintain semantics.
While constructing semantics-preserving continuous noise in a high-dimensional space proves to be non-trivial, state-of-the-art NMT models are currently based on adversarial examples of discrete noise.
For instance, \citet{Cheng:19} generate adversarial sentences using discrete word replacements in both the source and target, guided by the NMT loss. This approach achieves significant improvements over the Transformer on several standard NMT benchmarks.
Despite this promising result, we find that the generated adversarial sentences are unnatural, and, as we will show, suboptimal for learning robust NMT models.


In this paper, we propose {\em \fancyname}, a new adversarial augmentation technique for sequence-to-sequence learning. We introduce a novel vicinity distribution to describe the space of adversarial examples centered around each training example. Unlike prior work~\cite{Cheng:19}, we first generate adversarial sentences in the discrete data space and then sample \emph{virtual} adversarial sentences from the vicinity distribution according to their interpolated embeddings. Our intuition is that the introduced vicinity distribution may increase the sample diversity for adversarial sentences. Our idea is partially inspired by \mixup~\cite{Zhang:18}, a technique for data augmentation in computer vision, and we also use a similar vicinity distribution as in \mixup~ to augment the authentic training data.
Our {\em \fancyname} approach finally trains on the embeddings sampled from the above two vicinity distributions. As a result, we augment the training using virtual sentences in the feature space as opposed to in the data space. The novelty of our paper is the new vicinity distribution for adversarial examples and the augmentation algorithm for sequence-to-sequence learning.

Extensive experimental results on three translation benchmarks (NIST Chinese-English, IWSLT English-French, and WMT English-German) show that our approach achieves significant improvements of up to $4.9$ BLEU points over the Transformer~\cite{Vaswani:17}, outperforming the former state-of-the-art in adversarial learning \cite{Cheng:19} by up to $3.3$ BLEU points. When compared with widely-used data augmentation methods~\cite{Sennrich:16b, Edunov:18}, we find that our approach yields better performance even without using extra corpora.
We conduct ablation studies to gain further insights into which parts of our approach matter most. In summary, our contributions are as follows:
\begin{enumerate}
    \item We propose to sample adversarial examples from a new vicinity distribution and utilize their embeddings, instead of their data points, to augment the model training.
    \item We design an effective augmentation algorithm for learning sequence-to-sequence NMT models via mini-batches. 
    \item Our approach achieves significant improvements over the Transformer and prior state-of-the-art models on three translation benchmarks.
\end{enumerate}

\section{Background}

\paragraph{Neural Machine Translation.}
Generally, NMT ~\cite{Bahdanau:15, Gehring:17, Vaswani:17} models the translation probability $P(\mathbf{y}|\mathbf{x};\bm{\theta})$ based on the encoder-decoder paradigm where $\mathbf{x}$ is a source-language sentence, $\mathbf{y}$ is a target-language sentence, and $\bm{\theta}$ is a set of model parameters. 
The decoder in the NMT model acts as a conditional language model that operates on a shifted copy of $\mathbf{y}$, i.e., $\langle sos \rangle, y_{0},...,y_{|\mathbf{y}|-1}$ where $\langle sos \rangle$ is a start symbol of a sentence and representations of $\mathbf{x}$ learned by the encoder.
For clarity, we use $\ex \in \mathbb{R}^{d \times \vert \mathbf{x} \vert}$ to denote the feature vectors (or word embeddings) of the sentence $\mathbf{x}$ where $d$ is dimension size.

Given a parallel training corpus $\mathcal{S}$, the standard training objective for NMT is to minimize the empirical risk:
\begin{eqnarray}
\mathcal{L}_{clean}(\bm{\theta}) = \mathop{\mathbb{E}}\limits_{P_{\delta}(\bx, \by)} \lbrack \ell(f(\ex, \ey;\bm{\theta}), \ddot{\by}) \rbrack,
\label{eq:loss_clean}
\end{eqnarray}
where $f(\ex, \ey;\bm{\theta})$ is a sequence of model predictions $f_{j}(\ex, \ey;\bm{\theta}) = P(y|\mathbf{y}_{< j}, \mathbf{x};\bm{\theta}) $ at position $j$, and $\ddot{\by}$ is a sequence of one-hot label vectors for $\by$ (with label smoothing in the Transformer). $\ell$ is the cross entropy loss. The expectation of the loss function is summed over the empirical distribution $P_{\delta}(\bx,\by)$ of the training corpus:
\begin{eqnarray}
P_{\delta}(\bx, \by)  = \frac{1}{\vert S \vert}{\sum_{(\bxp, \byp) \in \mathcal{S}} \delta (\bx=\bxp, \by=\byp)},
\end{eqnarray}
where $\delta$ denotes the Dirac delta function.
 
\paragraph{Generating Adversarial Examples for NMT.} 
To improve NMT's robustness to small perturbations in the input sentences, \citet{Cheng:19} incorporate adversarial examples into the NMT model training. These adversarial sentences $\bx^{\prime}$ are generated by applying small perturbations that are jointly learned together with the NMT model:
\begin{equation}
    \hbx = \argmax_{\hbx : \mathcal{R}(\hbx, \bx) \leq \epsilon} \ell(f(e(\hat{\bx}), \ey;\bm{\theta}), \ddot{\by}), \label{eq:adv_space}
\end{equation}
where $\mathcal{R}(\hbx, \bx)$ captures the degree of semantic similarity and
$\epsilon$ is an upper bound on the semantic distance between the adversarial example and the original example. Ideally, the adversarial sentences convey only barely perceptible differences to the original input sentence yet result in dramatic distortions of the model output.

\citet{Cheng:19} propose the \emph{AdvGen} algorithm, which greedily replaces words with their top $k$ most probable alternatives, using the gradients of their word embeddings. 
Adversarial examples are designed to both attack and defend the NMT model. On the encoder side, an adversarial sentence $\hbx$ is constructed from the original input $\bx$ to attack the NMT model. To defend against adversarial perturbations in the source input $\hbx$, they use the \emph{AdvGen} algorithm to find an adversarial target input $\hby$ from the decoder input $\by$.
For notational convenience, let $\pi$ denote this algorithm, the adversarial example $\hat{\mathbf{s}}$ is stochastically induced by $\pi$ as $\hat{\mathbf{s}} \leftarrow \pi(\mathbf{s};\mathbf{x}, \mathbf{y}, \xi)$ where $\xi$ is the set of parameters used in $\pi$ including the NMT model parameters $\bm{\theta}$. 
For a detailed definition of $\xi$, we refer to \cite{Cheng:19}.
Hence, the set of adversarial examples originating from $(\bx,\by) \in \S$, namely $A_{(\mathbf{x}, \mathbf{y})}$, can be written as:
\begin{eqnarray}
A_{(\mathbf{x}, \mathbf{y})}=
\{
(\hbx, \hby)|
\hbx \leftarrow \pi(\bx;\bx, \by, \xi_{src}) \nonumber, \\
\hby \leftarrow \pi(\by;\hbx, \by, \xi_{tgt}) \}, \label{eq:adv_set}
\end{eqnarray}
where $\xi_{src}$ and $\xi_{tgt}$ are separate parameters for generating $\hbx$ and $\hby$, respectively. 
Finally, the robustness loss $\mathcal{L}_{robust}$ is computed on $A_{(\bx,\by)}$ with the loss $\ell(f(e(\hbx), e(\hby);\bm{\theta}), \ddot{\by})$, and is used together with $\mathcal{L}_{clean}$ to train the NMT model.

\paragraph{Data Mixup.}
In image classification, the \mixup data augmentation technique involves training on linear interpolations of randomly sampled pairs of examples~\cite{Zhang:18}.
Given a pair of images $(\bxp, \byp)$ and $(\bxpp, \bypp)$, where $\bxp$ denotes the RGB pixels of the input image and $\byp$ is its one-hot label, \mixup minimizes the sample loss from a vicinity distribution~\cite{chapelle2001vicinal} $P_v(\tilde{\mathbf{x}}, \tilde{\mathbf{y}}) $ defined in the RGB-pixel (label) space:
\begin{eqnarray}
\tbx = \lambda \bxp + (1 - \lambda) \bxpp, \\
\tby = \lambda \byp + (1 - \lambda) \bypp.
\end{eqnarray}
$\lambda$ is drawn from a Beta distribution $\text{Beta}(\alpha, \alpha)$ controlled by the hyperparameter $\alpha$.
When $\alpha \rightarrow 0$, $(\tbx, \tby)$ is close to any one of the images $(\bxp, \byp)$ and $(\bxpp, \bypp)$.
Conversely, $(\tbx, \tby)$ approaches the middle interpolation point between them when $\alpha \rightarrow  + \infty$. 
The neural networks $g$ parameterized by $\psi$ can be trained over the mixed images $(\tbx, \tby)$ with the loss function
$\mathcal{L}_{mixup}(\bm{\theta}) = \ell(g(\tilde{\bx};\bm{\psi}), \tilde{\by})$. In practice, the image pair is randomly sampled from the same mini-batch.

\begin{figure}[!t]
\centering
\includegraphics[width=0.47\textwidth]{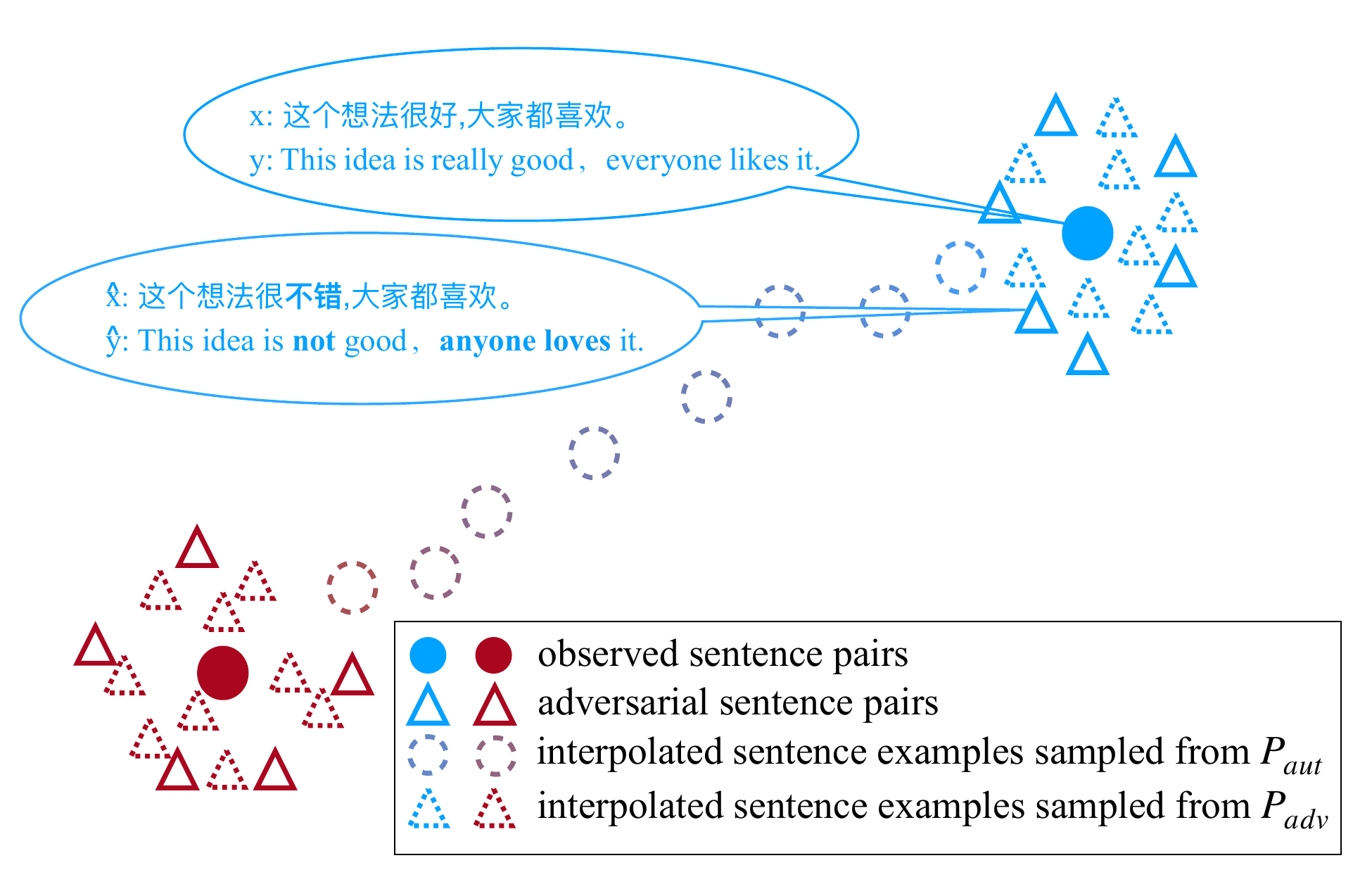} 
\caption{Illustration of training examples sampled from vicinity distributions of $P_{adv}$ and $P_{aut}$. Solid circles are observed sentences in the training corpus $\mathcal{S}$. Solid triangles are adversarial sentences generated by replacing words in their corresponding observed sentences. 
Dashed points are virtual sentences obtained by interpolating the embeddings of the solid points. 
The dashed triangles define the data space of adversarial examples from $P_{adv}$. The circles (solid and dashed) constitute $P_{aut}$.} \label{figure:approach} 
\label{fig:train}
\end{figure}

\section{Approach: \fancyname}
In our approach {\em \fancyname}, the goal is to reinforce the model over virtual data points surrounding the observed examples in the training set.

We approximate the density of $P(\bx, \by)$ in the vicinities of the generated adversarial examples and observed training examples. To be specific, we design two vicinity distributions~\cite{chapelle2001vicinal} to estimate the joint distribution of $P(\bx, \by)$: $P_{adv}$ for the (dynamically generated) \emph{adversarial} examples and $P_{aut}$ for the (observed) \emph{authentic} examples in $\S$. Given the training set $\S$, we have:
\begin{align}
P_{adv}(\tbx, \tby) &=  \frac{1}{\cS} \sum_{ (\bx, \by) \in \S} \mu_{adv} (\tbx, \tby | A_{(\bx, \by)}), \\
P_{aut}(\tbx, \tby) &= \frac{1}{\cS} \sum_{ (\bx, \by) \in \S} \mu_{aut} (\tbx, \tby | \bx, \by),
\end{align}
where $A_{(\bx, \by)}$ is the set of adversarial examples originated from $(\bx, \by)$ defined in Eq.~\eqref{eq:adv_set}. We will discuss $\mu_{adv}$ and $\mu_{aut}$ in detail which define the probability functions, but first we give some high-level descriptions:
\begin{itemize}[leftmargin=*]
    \item $P_{adv}$ is a new vicinity distribution for virtual adversarial sentences of the same origin. It captures the intuition that the convex combination of adversarial sentences should have the same translation. It is the most important factor for improving the translation quality in our experiments.
    \item $P_{aut}$ is a distribution to improve the NMT's robustness by ``mixing up'' observed sentences of different origins. This distribution is similar to \mixup, but it is defined over linear interpolations of the sequence of word embeddings of the source and target sentences. Although $P_{aut}$ by itself yields marginal improvements, we find it is complementary to $P_{adv}$.
\end{itemize}

We train the NMT model on two vicinity distributions $P_{adv}$ and $P_{aut}$. Figure~\ref{figure:approach} illustrates examples sampled from them.
As shown, a solid circle stands for an observed training example (\ie a sentence-pair) in $\S$ and a solid triangle denotes an adversarial example in $A_{(\bx, \by)}$. For $P_{adv}$, we construct virtual \emph{adversarial} examples (dashed triangles) to amend the sample diversity by interpolating the word embeddings of solid triangles. Likewise, we interpolate the word embeddings of solid circles to model $P_{aut}$ for the (observed) \emph{authentic} examples. This results in the dashed circles in Figure~\ref{figure:approach}.


Unlike prior works on vicinal risk minimization~\cite{chapelle2001vicinal,Zhang:18}, we do not directly observe the virtual sentences in $P_{adv}$ or $P_{aut}$. This also distinguishes us from~\citet{Cheng:19}, who generate actual adversarial sentences in the discrete word space. In the remainder of this section, we will discuss the definition of $P_{adv}$ and $P_{aut}$ and how to optimize the translation loss over virtual sentences via mini-batch training.

\subsection{$P_{adv}$ for Adversarial Data}

To compute $\mu_{adv}$, we employ $\pi$ similar as in ~\cite{Cheng:19} to generate 
an adversarial example set $A_{(\bx, \by)}$ from each instance $(\bx,\by) \in \S$ (see~Eq.~\eqref{eq:adv_set}). Let $(\bxp, \byp)$ and $(\bxpp, \bypp)$ be two examples randomly sampled from $A_{(\bx, \by)}$. We align the two sequences by padding tokens to the end of the shorter sentence. Note that this operation aims for a general case (particularly for $P_{aut}$) although the lengths of $\byp$ and $\bypp$ in $A_{(\bx, \by)}$ are same.
To obtain $\tex = [e(\tilde{x}_{1}), \ldots, e(\tilde{x}_{|\tbx|})]$, we apply the convex combination $m_{\lambda}(\bxp, \bxpp)$ over the aligned word embeddings, which is:
\begin{equation}
e(\tilde{x}_{i}) \!=\! \lambda e(x^{\prime}_i) + (1 - \lambda) e(x^{\prime\prime}_{i}),  \forall i \in [1, \vert \tbx \vert ], \label{eq:mix_src}
\end{equation}
where $\lambda \sim \text{Beta}(\alpha, \alpha)$. We use $m_{\lambda}(\cdot, \cdot)$ for the interpolation. Similarly, $e(\tby)$ can also be obtained with $m_{\lambda}(\byp, \bypp)$. 

All adversarial examples in $A_{(\bx, \by)}$ are supposed to be translated into the same target sentence, and the convex combination still lies in space of the adversarial search ball defined in Eq.~\eqref{eq:adv_space}. As a result, all virtual sentence pairs $(\tbx, \tby) \in A_{(\bx, \by)}$ of the same origin can be fed into NMT models as source and target inputs which share the same soft target label for $(\bx, \by)$.

$\mu_{adv}$ in $P_{adv}$ can be calculated from:
\begin{align}
\mu_{adv} (\tbx, \tby | \Axy)=
\frac{1}{\vert \Axy \vert ^2} \quad \nonumber \\
\!\!\!\!\!\sum_{(\bxp, \byp) \in \Axy}\!\! \sum_{(\bxpp, \bypp) \in \Axy}   
\mathop{\mathbb{E}}\limits_{\lambda} [\delta(\tex=m_{\lambda}(\bxp, \bxpp), \nonumber \\
\tey=m_{\lambda}(\byp, \bypp)]. \label{mu_adv}
\end{align}

Hence, the translation loss on vicinal adversarial examples $\mathcal{L}_{adv}(\bm{\theta})$ can be integrated over $P_{adv}$ as:
\begin{eqnarray}
\mathcal{L}_{adv}(\bm{\theta})\!=\!\!\! \mathop{\mathbb{E}}\limits_{P_{adv}{(\tbx,\tby)}} \lbrack \ell(f(\tex, \tey;\bm{\theta}), \bm{\omega}) \rbrack,
\label{eq:train_loss_adv}
\end{eqnarray}
where $\bm{\omega}$ is a sequence of output distributions (denoted as a sequence of label vectors, \eg $\ddot{\by}$) as the soft target for the sentence $\mathbf{y}$. 

We employ two useful techniques in computing the loss $\mathcal{L}_{adv}$ in Eq.~\eqref{eq:train_loss_adv}. First, we minimize the KL-divergence between the model predictions at the word level:
\begin{equation}
\sum_{j=1} ^{\vert \by \vert} D_{KL}(f_{j}(e(\bx), e(\by);\hat{\bm{\theta}}) || f_{j}(\tex, \tey ;\bm{\theta})), \label{eq:kl_adv}
\end{equation}
where $\hat{\bm{\theta}}$ means a fixed copy of the current parameter set and no gradients are back-propagated through it. Removing constant values from Eq.~\eqref{eq:kl_adv} yields an equivalent solution of:
\begin{align}
&\ell(f(\tex, \tey ;\bm{\theta}),\bm{\omega}) \nonumber \\
=&\ell(f(\tex, \tey ;\bm{\theta}),f(\ex, \ey;\hat{\bm{\theta}})).
\label{eq:loss_adv}
\end{align}
Eq.~\eqref{eq:loss_adv} indicates that $f(e(\bx), e(\by); \hat{\bm{\theta}})$ can be used as the soft target $\bm{\omega}$ in Eq.~\eqref{eq:train_loss_adv} for virtual adversarial example $(\tbx, \tby)$.
KL-divergence enforces the model on virtual adversarial examples to indirectly learn from the soft target of the observed examples over large vocabularies. This justifies the use of $\bm{\omega}$ in Eq.~\eqref{eq:train_loss_adv} and turns out to be more effective than directly learning from the ground-truth label.

Besides, Eq.~\eqref{eq:train_loss_adv} needs to enumerate numerous pairs of adversarial examples in $\Axy$ while in practice we only sample a pair at a time inside each mini-batch for training efficiency.
We hence employ curriculum learning to do the importance sampling. To do so, we re-normalize the translation loss and employ a curriculum from \cite{jiang2017mentornet} to encourage the model to focus on the difficult training examples. 

Formally, for a mini-batch of the training losses $\mathbf{L}=\{\ell_{i} \}_{i=1}^m$, we re-weigh the batch loss using:
\begin{eqnarray}
\mathbf{L} = \frac{1}{\sum_{i=1}^m I(\ell_{i} > \eta) } \sum_{i=1}^m I(\ell_{i} > \eta) \ell_{i},
\label{eq:loss_reweight}
\end{eqnarray}
where $I(\cdot)$ is an indicator function and $\eta$ is set by a moving average tracking the $p$-th percentile of the example losses of every mini-batch. In our experiments, we set the $p$-th percentile to be $100 \times (1-r_{t})$ for the training iteration $t$, 
and gradually anneal $r_t$ using $r_{t} = 0.5^{t/\beta}$, where $\beta$ is the hyperparameter.
\subsection{$P_{aut}$ for Authentic Data}
We define the $\mu_{aut}$  in the vicinity distribution $P_{aut}$ for authentic examples as follows:
\begin{eqnarray}
\mu_{aut}(\tbx, \tby | \bx, \by) = \frac{1}{\cS} \sum\limits_{(\bxp, \byp) \in \S} \mathop{\mathbb{E}}\limits_{\lambda}[ \nonumber \\ \delta(\tex=m_{\lambda}(\bx, \bxp), 
\tey=m_{\lambda}(\by, \byp), \nonumber \\
\tilde{\bm{\omega}}=m_{\lambda}(\bm{\omega}, \bm{\omega^{\prime}}))].
\label{eq:mu_aut}
\end{eqnarray}

The translation loss on authentic data is integrated over all examples of the vicinity distribution $P_{aut}$:
\begin{eqnarray}
\mathcal{L}_{aut}(\bm{\theta}) =\!\!\! \mathop{\mathbb{E}}\limits_{P_{aut}{(\tbx,\tby)}} \lbrack \ell(f(\tex, \tey;\bm{\theta}), \tilde{\bm{\omega}}) \rbrack.
\label{eq:loss_aut}
\end{eqnarray}
In our experiments, we select the value of $\lambda$ in Eq.~\eqref{eq:mu_aut} twice for every $(\bx, \by)$: (1) a constant $1.0$ and (2) a sample from the Beta distribution.
The former is equivalent to sampling from the empirical distribution $P_{\delta}$ whereas the latter is similar to applying \mixup in the embedding space of the sequence model. In other words, $\mathcal{L}_{aut}(\bm{\theta})$ equals the sum of two translation losses, $\mathcal{L}_{clean}(\bm{\theta})$ computed on the original training examples when $\lambda$ is $1.0$ and $\mathcal{L}_{mixup}(\bm{\theta})$ computed on virtual examples when $\lambda$ is sampled from a Beta distribution. Accordingly, when  $\lambda$ is $1.0$ we set $\tilde{\bm{\omega}}$ to be the interpolation of the sequences of one-hot label vectors for $\by$ and $\byp$, \ie $\bm{\omega} = \ddot{\by}$ and $\bm{\omega^{\prime}} = \ddot{\mathbf{y}}^{\prime}$.
Otherwise $\tilde{\bm{\omega}}$ is the interpolation of model output vectors of $(\bx, \by)$ and $(\bxp, \byp)$, that is, $\bm{\omega}=f(\ex, \ey;\hat{\bm{\theta}})$ and $\bm{\omega^{\prime}}=f(e(\bxp), e(\byp);\hat{\bm{\theta}})$.
\begin{algorithm}[!t]
\small
\SetAlgoLined
\LinesNumbered
\KwIn{A batch of source and target sentences ($\mathbf{X}, \mathbf{Y}$); the selection ratio $r_{t}$; the hyperparameter $\alpha$.}
\KwOut{Mini-batch losses $\mathcal{L}_{adv}$ and $\mathcal{L}_{aut}$ }
\SetKwFunction{algo}{\fancyname}
\SetKwProg{Fn}{Function}{:}{}
\Fn{\algo{$\mathbf{X}, \mathbf{Y}$}}{
  \ForEach{ $(\bx, \by)$ $\in$ ($\mathbf{X}, \mathbf{Y}$) }{
    
    $\bm{\omega} \leftarrow f(\ex, \ey; \hat{\bm{\theta}})$\;
    Sample two adversarial examples ($\bxp, \byp$) and ($\bxpp, \bypp$) from $\Axy$ by Eq.~\eqref{eq:adv_set} \;
    $\lambda \leftarrow \text{Beta}(\alpha, \alpha)$ \;
    $e(\tbx) \leftarrow m_{\lambda}(\bxp, \bxpp)$, $e(\tby) \leftarrow m_{\lambda}(\byp, \bypp)$\;
    $\ell_{i} \leftarrow \ell(f(e(\tbx), e(\tby);\bm{\theta}), \bm{\omega})$ \;
    }
  Compute $\mathcal{L}_{adv}$ using $r_{t}$ and $\{ \ell_{i}\}$  by Eq. ~\eqref{eq:loss_reweight} \;
   ($\mathbf{X}^{\prime}, \mathbf{Y}^{\prime}$) $\leftarrow$ Shuffle ($\mathbf{X}, \mathbf{Y}$) \;
    \ForEach{ $(\bx, \by, \bxp, \byp)$ $\in$ ($\mathbf{X}, \mathbf{Y},\mathbf{X}^{\prime}, \mathbf{Y}^{\prime}$) }{
    
    $\bm{\omega} \leftarrow f(\ex, \ey; \hat{\bm{\theta}})$\;
    $\bm{\omega}^{\prime} \leftarrow f(e(\bxp), e(\byp); \hat{\bm{\theta}})$\;
    $\lambda \leftarrow \text{Beta}(\alpha, \alpha)$ \;
    $e(\tbx) \leftarrow m_{\lambda}(\bx, \bxp)$, $e(\tby) \leftarrow m_{\lambda}(\by, \byp)$ \;
    $\tilde{\bm{\omega}} \leftarrow m_{\lambda}(\bm{\omega}, \bm{\omega}^{\prime})$ \;
    $\ell_{i} \leftarrow \ell(f(e(\tbx), e(\tby);\bm{\theta}), \tilde{\bm{\omega}})$  + 
    $\ell(f(e(\bx), e(\by);\bm{\theta}), \ddot{\by})$ \;
  }
   Compute $\mathcal{L}_{aut}$ by averaging $\{ \ell_{i} \}$ \;
 \KwRet $\mathcal{L}_{adv}$, $\mathcal{L}_{aut}$ }
 \caption{Proposed {\em \fancyname} function.} \label{algo2}
\end{algorithm}
\subsection{Training}
Finally, the training objective in our {\em \fancyname} is a combination of the two losses:
\begin{eqnarray}
\bm{\theta}^* = \argmin\limits_{\bm{\theta}} \lbrace {\mathcal{L}_{aut}(\bm{\theta}) + \mathcal{L}_{adv}(\bm{\theta})} \rbrace. \label{eq:training_loss}
\end{eqnarray}
Here, we omit two bidirectional language model losses for simplicity, which are used to recommend word candidates to maintain semantic similarities \cite{Cheng:19}.

In practice, we need to compute the loss via mini-batch training. For the $P_{aut}$, we follow the pair sampling inside each mini-batch in \mixup. It can avoid padding too much tokens because sentences of similar lengths are grouped within a mini-batch \cite{Vaswani:17}. For the $P_{adv}$, we sample a pair of examples from $A_{(\bx, \by)}$ for each $(\bx, \by)$ and cover the distribution over multiple training epochs. The entire procedure to calculate the translation losses, $\mathcal{L}_{adv}(\bm{\theta})$ and $\mathcal{L}_{aut}(\bm{\theta})$, is presented in Algorithm \ref{algo2}. 
In a nutshell, for each batch of training examples, we firstly sample virtual examples from $P_{adv}$ and $P_{aut}$ by interpolating the embeddings of the adversarial or authentic training examples. Then we calculate the translation loss using their interpolated embeddings. 

\section{Experiments}

\begin{table*}[!t]
\centering
\begin{tabular}{l|l|l|lllll}
\hline
{Method}  &Loss Config. &{MT06} &{MT02} & {MT03} & {MT04} & {MT05} &{MT08}\\
\hline \hline  
\multicolumn{1}{c|}{\newcite{Vaswani:17}} &$\mathcal{L}_{clean}$ &44.57 &45.49 &44.55 &46.20 &44.96 &35.11  \\
\multicolumn{1}{c|}{\newcite{Miyato:17}} & -&45.28 &45.95 &44.68 &45.99 &45.32 &35.84 \\
\multicolumn{1}{c|}{\newcite{Sano:19}} & -&45.75 &46.37 &45.02 &46.49 &45.88 &35.90\\
\multicolumn{1}{c|}{\newcite{Cheng:19}} &- &46.95 &47.06 &46.48 &47.39 &46.58 &37.38 \\
\multicolumn{1}{c|}{\newcite{Sennrich:16b}*} &- &46.39 &47.31 &47.10 &47.81 &45.69  &36.43\\
\multicolumn{1}{c|}{\newcite{Edunov:18}*} &- &46.20 &47.78 &46.93 &47.80 &46.81 &36.79 \\
\hline
\multirow{4}{*}{Ours} &$\mathcal{L}_{mixup}$ &45.12 &46.32 &44.81 &46.61 &46.08 &36.00\\
  &$\mathcal{L}_{aut}$ &46.73 &46.79 &46.13 &47.54 &46.88 &37.21\\
 &$\mathcal{L}_{clean} + \mathcal{L}_{adv}$ &47.89 &48.53 &\textbf{48.73} &48.60 &48.76 &39.03\\
  &$\mathcal{L}_{aut} + \mathcal{L}_{adv}$ &\textbf{49.26} &\textbf{49.03} &47.96 &\textbf{48.86} &\textbf{49.88} &\textbf{39.63}\\
\hline
Ours* &$\mathcal{L}_{aut} + \mathcal{L}_{adv}$ &\textbf{49.98} &\textbf{50.34} &\textbf{49.81} &\textbf{50.61} &\textbf{50.72} &\textbf{40.45} \\
\hline
\end{tabular}
\caption{Baseline comparison on NIST Chinese-English translation. * indicates the model uses extra corpora and - means not elaborating on its training loss.}
\label{table:comparison_zhen}
\end{table*}

\subsection{Setup}
We verify our approach on translation tasks for three language pairs: Chinese-English, English-French, and English-German. 
The performance is evaluated with the 4-gram BLEU score \cite{Papineni:02} calculated by the {\em multi-bleu.perl} script. We report case-sensitive tokenized BLEU scores for English-French and  English-German, and case-insensitive tokenized BLEU scores for Chinese-English.
Note that all reported BLEU scores in our approach are from a single model rather than averaging multiple models \cite{Vaswani:17}.

For the Chinese-English translation task, the training set is the LDC corpus consisting of 1.2M sentence pairs. 
The NIST 2006 dataset is used as the validation set, and NIST 02, 03, 04, 05, 08 are used as the test sets.
We apply byte-pair encoding (BPE) \cite{Sennrich:16a} with 60K merge operations to build two vocabularies comprising 46K Chinese sub-words and 30K English sub-words.
We use the IWSLT 2016 corpus for English-French translation. The training corpus with 0.23M sentence pairs is preprocessed with the BPE script with 20K joint operations. The validation set is test2012 and the test sets are test2013 and test2014. For English-German translation, we use the WMT14 corpus consisting of 4.5M sentence pairs. The validation set is newstest2013 whereas the test set is newstest2014. We build a shared vocabulary of 32K sub-words using the BPE script.

We implement our approach on top of the Transformer \cite{Vaswani:17}. The size of the hidden unit is $512$ and the other hyperparameters are set following their default settings. 
There are three important hyperparameters in our approach, $\alpha$ in the Beta distribution and the word replacement ratio of $\gamma_{src} \in \xi_{src}$, and $\gamma_{tgt} \in \xi_{tgt}$ detailed in Eq.~\eqref{eq:adv_set}. Note that $\gamma_{src}$ and $\gamma_{tgt}$ are not new hyperparameters but inherited from~\cite{Cheng:19}. We tune these hyperameters on the validation set via a grid search, \ie $\alpha \in \{ 0.2, 0.4, 4, 8, 32\}$, $\gamma_{src} \in \{ 0.10, 0.15, 0.25\}$ and $\gamma_{tgt} \in \{ 0.10, 0.15, 0.30, 0.5\}$. For the \mixup loss $\mathcal{L}_{mixup}$, $\alpha$ is fixed to $0.2$. For the loss $\mathcal{L}_{aut}$ and $L_{adv}$, the optimal value of $\alpha$ is $8.0$. The optimal values of $(\gamma_{src}, \gamma_{tgt})$ are found to be $(0.25, 0.50)$, $(0.15, 0.30)$ and $(0.15, 0.15)$ for Chinese-English, English-French and English-German, respectively, while it is set to $(0.10, 0.10)$ only for back-translated sentence pairs. $\beta$ in Eq.~\eqref{eq:loss_reweight} is set to 250K, 100K, 1M for Chinese-English, English-French and English-German.
Unlike \citet{Cheng:19}, we remove the learning of target language models to speed up the training. For each training batch, we introduce a batch of augmented adversarial examples and a batch of augmented authentic examples, which costs twice the vanilla training. For constructing adversarial examples, we solely compute the gradients for word embeddings which takes little time. After summing up the time of all steps, our total training time is about 3.3 times the vanilla training.

\begin{table*}[!t]
\centering
\begin{tabular}{l|l|ll|ll}
\hline                                      
\multirow{2}{*}{Method} &\multirow{2}{*}{Loss Config.}  &\multicolumn{2}{c|}{English-French} &\multicolumn{2}{c}{English-German} \\
\cline{3-6}
& &test2013 &test2014 &newstest13 &newstest14\\
\hline \hline
\multicolumn{1}{c|}{\newcite{Vaswani:17}} & $\mathcal{L}_{clean}$  &40.78  &37.57 &25.80 &27.30 \\
\multicolumn{1}{c|}{\newcite{Sano:19}} & $-$  &41.67  &38.72 &25.97 &27.46 \\
\multicolumn{1}{c|}{\newcite{Cheng:19}} & $-$  &41.76  &39.46 &26.34 &28.34 \\
\hline 
\multirow{3}{*}{Ours} &$\mathcal{L}_{mixup}$  &40.78  &38.11 &26.28 &28.08 \\

&$\mathcal{L}_{aut}$  &41.49  &38.74 &26.33 &28.58 \\
  &$\mathcal{L}_{aut} + \mathcal{L}_{adv}$  & \textbf{43.03}  &\textbf{40.91} &\textbf{27.20} &\textbf{29.57}\\ 
\hline
\end{tabular}
\caption{Results on IWSLT16 English-French and WMT14 English-German translation.} \label{table:en-de}
\label{table:comparison_ende}
\end{table*}

\subsection{Main Results}

\textbf{Chinese-English Translation.}
\autoref{table:comparison_zhen} shows results on the Chinese-English translation task, in comparison with the following six baseline methods. For a fair comparison, we implement all these methods using the Transformer backbone or report results from those papers on the same corpora.
\begin{enumerate}[itemsep=0pt,parsep=0pt,leftmargin=3ex]
\item The seminal Transformer model for NMT~\cite{Vaswani:17}. 
\item Following \citet{Miyato:17}, we use adversarial learning to add continuous gradient-based perturbations to source word embeddings and extend it to the Transformer model.
\item \citet{Sano:19} leverage \citet{Miyato:17}'s idea into NMT by incorporating gradient-based perturbations to both source and target word embeddings and optimize the model with adversarial training. 
\item \citet{Cheng:19} generate discrete adversarial examples guided by the gradients of word embeddings. Adversarial examples are used to both attack and defend the NMT model.
\item \citet{Sennrich:16b} translate monolingual corpora using an inverse NMT model and then augment the training data with them.
\item Based on \citet{Sennrich:16b}, \citet{Edunov:18} propose three improved methods to generate back-translated data, which are {\em sampling}, {\em top10} and {\em beam+noise}. Among those, we choose {\em beam+noise} as our baseline method, which can be regarded as an approach to incorporating noise into data.
\end{enumerate}

We first verify the importance of different translation losses in our approach.
We find that both $\mathcal{L}_{aut}$ and $\mathcal{L}_{adv}$ are useful in improving the Transformer model. $\mathcal{L}_{adv}$ is more important and yields a significant improvement when combined with the standard empirical loss $\mathcal{L}_{clean}$ (cf.~Eq.~\eqref{eq:loss_clean}). These results validate the effectiveness of augmenting with virtual adversarial examples. When we use both $\mathcal{L}_{aut}$ and $\mathcal{L}_{adv}$ to train the model, we obtain the best performance (up to 4.92 BLEU points on MT05). We also compare with the \mixup loss. However, $\mathcal{L}_{mixup}$ is only slightly better than the standard empirical loss $\mathcal{L}_{clean}$.

Compared with the baseline methods without using extra corpora, our approach shows significant improvements over the state-of-the-art models. In particular, the superiority of $\mathcal{L}_{clean}+\mathcal{L}_{adv}$ over both \citet{Cheng:19} and \citet{Sano:19} verifies that we propose a more effective method to address adversarial examples in NMT. We also directly incorporate two adversarial examples to NMT models without interpolating their embeddings, but we do not observe any further gain over \citet{Cheng:19}. This substantiates the superior performance of our approach on the standard data sets.

To compare with the approaches using extra monolingual corpora, we sample 1.25M English sentences from the Xinhua portion of the GIGAWORD corpus and list our performance in the last row of Table~\ref{table:comparison_zhen}.
When the back-translated corpus is incorporated, our approach yields further improvements, suggesting our approach complements the back-translation approaches.

\noindent\textbf{English-French and English-German Translation.} Table \ref{table:en-de} shows the comparison with the Transformer model \cite{Vaswani:17}, \citet{Sano:19} and \citet{Cheng:19} on English-French and English-German translation tasks.
Our approach consistently outperforms all three baseline methods, yielding significant $3.34$ and $2.27$ BLEU point gains over the Transformer on the English-French and English-German translation tasks, respectively.
We also conduct similar ablation studies on the translation loss.
We still find that the combination of $\mathcal{L}_{adv}$ abd $\mathcal{L}_{aut}$ performs the best, which is consistent with the findings in the Chinese-English translation task.
The substantial gains on these two translation tasks suggest the potential applicability of our approach to more language pairs.

\begin{table*}[!t]
\centering
\begin{tabular}{c|l}
\hline
  Input & \begin{CJK*}{UTF8}{gbsn}{但（{\color{red}{但是}}）协议执行过程一波三折，致使和平进程一再受挫}\end{CJK*}\\
 \hline
 Reference & however, implementation of the deals has witnessed ups and downs, resulting \\ & in continuous setbacks in the peace process \\
\hline \hline
\citeauthor{Vaswani:17} & however, the process of implementing the agreement was full of twists and \\ 
on Input  &turns, with the result that the peace process suffered setbacks again and again.\\
\hline

on Noisy Input & the process of the agreement has caused repeated setbacks to the peace process.       \\
\hline \hline
Ours & however, the process of implementing the agreement experienced twists and \\
on Input &turns, resulting in repeated setbacks in the peace process.\\
\hline
on Noisy Input & however, the process of implementing the agreement experienced twists and\\
    &turns, resulting in repeated setbacks in the peace process. \\
\hline
\end{tabular}
\caption{Translation Examples of Transformer and our model for an input and its adversarial input.}
\label{table:robust_examples} 
\end{table*}

\begin{table}[!t]
\centering
\setlength\tabcolsep{3.9pt}
\begin{tabular}{l|l|l|l|l|l}
\hline
\multirow{2}{*}{Loss} &\multicolumn{5}{c}{$\alpha = $} \\
\cline{2-6}
&0.2 &0.4 &4 &8 & 32 \\

\hline \hline
$\mathcal{L}_{mixup}$ &45.28 &45.38 &45.64 &45.09 & -\\
{$\mathcal{L}_{aut}$} &45.95 &45.92 &46.70 &46.73 &46.54\\
{\!\tiny $\mathcal{L}_{clean}\!\! +\!\! \mathcal{L}_{adv}$\!} &47.06 &46.88  &47.60  &47.89 &47.81 \\
\hline
\end{tabular}
\caption{Effect of $\alpha$ on the Chinese-English validation set. ``-'' indicates that the model fails to converge.}
\label{table:alpha} 
\end{table}

\begin{table}[!t]
\centering
\setlength\tabcolsep{4.5pt}
\begin{tabular}{l|l|l|l|l}
\hline
Method &0.00 &0.05 &0.10 &0.15 \\
\hline \hline
\citeauthor{Vaswani:17} &44.59 &41.54 &38.84 &35.71\\
\citeauthor{Miyato:17} &45.11 &42.11 &39.39 &36.44\\
\citeauthor{Sano:19} &45.75 &44.04 &41.25 &38.78\\
\citeauthor{Cheng:19} &46.95 &44.20 &41.71 &39.89\\
\hline
Ours &\textbf{49.26} &\textbf{47.53} &\textbf{44.71} &\textbf{41.76}\\
\hline
\end{tabular}
\caption{Results on artificial noisy inputs. The column lists results for different noise fractions.}
\label{table:robust}
\end{table}

\subsection{Effect of $\alpha$}
The hyperparameter $\alpha$ controls the shape of the Beta distribution over interpolation weights. 
We study its effect on the validation set in \autoref{table:alpha}.
Notable differences occur when $\alpha < 1$ and $\alpha > 1$, this is because the Beta distribution show two different shapes with $\alpha = 1$ as a critical point. 
As we see, both $\mathcal{L}_{aut}$ and $\mathcal{L}_{adv}$ prefer a large $\alpha$ and perform better when $\alpha=8$. Recall that when $\alpha$ is large, $m_{\lambda}$ behaves similarly to a simple average function. 
In $\mathcal{L}_{mixup}$, $\alpha=4$ performs slightly better, and a large $\alpha=32$ will fail the model training. Although the result with $\alpha=4$ appears to be slightly better, it consumes more iterations to train the model to reach the convergence, \ie, $90$K for $\alpha=4$ vs. $20$K for $\alpha=0.2$. These indicate the differences between the proposed vicinity distributions and the one used in \mixup. 

\begin{figure}[!t]
\centering
\includegraphics[width=0.48\textwidth]{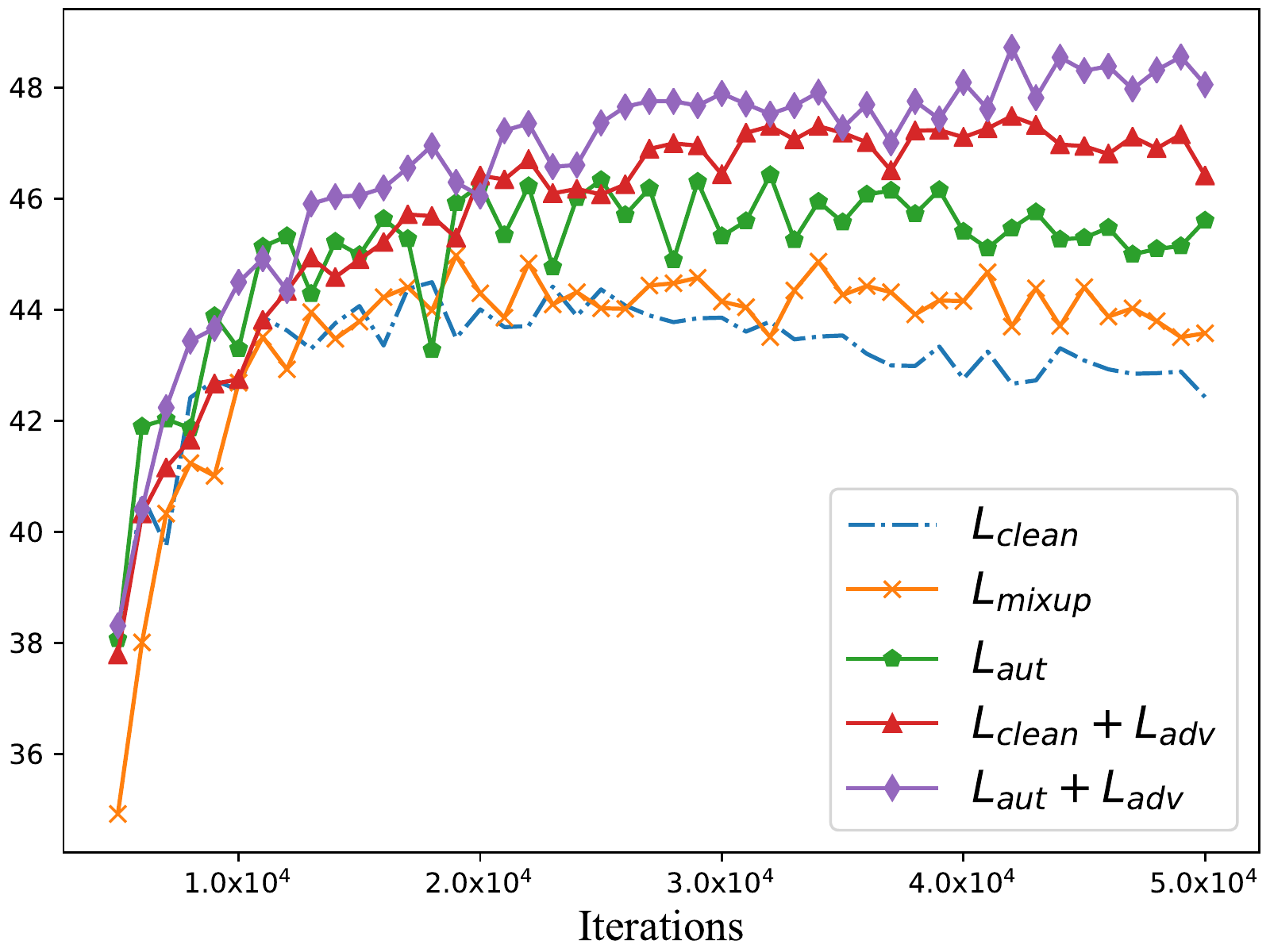} 
\caption{BLEU scores over iterations on the Chinese-English validation set.}
\label{fig:training_curve}
\end{figure}

\subsection{Robustness to Noisy Inputs and Overfitting}
To test robustness on noisy inputs, we follow \citet{Cheng:19} to construct a noisy data set by randomly replacing a word in each sentence of the standard validation set with a relevant alternative. The relevance between words is measured by the similarity of word embeddings. 100 noisy sentences are generated for each of them and then re-scored to pick the best one with a bidirectional language model. Table \ref{table:robust} shows the results on artificial noisy inputs with different noise levels. Our approach shows higher robustness over all baseline methods across all noise levels.

\autoref{fig:training_curve} shows the evolution of BLEU scores during training. For $\mathcal{L}_{clean}$, the BLEU score reaches its peak at about 20K iterations, and then the model starts overfitting. In comparison, all of the training losses proposed in this paper are capable of resisting overfitting: in fact, even after 100K iterations, no significant regression is observed (not shown in this figure).
At the same iteration, our results are consistently higher than both the empirical risk ($\mathcal{L}_{clean}$) and \mixup ($\mathcal{L}_{mixup}$).

As shown in Table \ref{table:robust_examples}, the baseline yields an incorrect translation possibly because the word “danshi(\begin{CJK*}{UTF8}{gbsn}{但是}\end{CJK*})” seldom occurs in this context in our training data. In contrast, our model incorporates embeddings of virtual sentences that contain “danshi(\begin{CJK*}{UTF8}{gbsn}{但是}\end{CJK*})” or its synonym “dan(\begin{CJK*}{UTF8}{gbsn}{但}\end{CJK*})”. This encourages our model to learn to push their embeddings closer during training, and make our model more robust to small perturbations in real sentences.

\section{Related Work}
\paragraph{Data Augmentation.}
Data augmentation is an effective method to improve machine translation performance.
Existing methods in NMT may be divided into two categories, based upon extra corpora \cite{Sennrich:16b,Cheng:16,Zhang:16,Edunov:18} or original parallel corpora \cite{Fadaee:17,Wang:18,Cheng:19}. Recently, \mixup~\cite{Zhang:18} has become a popular data augmentation technique for semi-supervised learning~\cite{Berthelot:19} and overcoming real-world noisy data~\cite{Jiang:19}. Unlike prior works, we introduce a new method to augment the representations of the adversarial examples in sequence-to-sequence training of the NMT model.
Even without extra monolingual corpora, our approach substantially outperforms the widely-used back-translation methods~\cite{Sennrich:16b, Edunov:18}. Furthermore, we can obtain even better performance by including additional monolingual corpora.

\paragraph{Robust Neural Machine Translation.}
It is well known that neural networks are sensitive to noisy inputs \cite{Szegedy:14, Goodfellow:14}, and neural machine translation is no exception.
Thus improving the robustness of NMT models has become a popular research topic~\cite[e.g.,][]{Belinkov:17,Sperber:17, Ebrahimi:18b, Cheng:18, Cheng:19, Karpukhin:19, Li:19}. 
Many of these studies focus on augmenting the training data to improve robustness, especially with adversarial examples \cite{Ebrahimi:18b, Cheng:19, Karpukhin:19, Michel:19}. Others also tried to deal with this issue by finding better input representations \cite{Durrani:19}, adding adversarial regularization \cite{Sano:19} and so on. In contrast to those studies, we propose the vicinity distribution defined in a smooth space by interpolating discrete adversarial examples.
Experimental results show substantial improvements on both clean and noisy inputs.

\section{Conclusion}
We have presented an approach to augment the training data of NMT models by introducing a new vicinity distribution defined over the interpolated embeddings of adversarial examples. To further improve the translation quality, we also incorporate an existing vicinity distribution, similar to \mixup for observed examples in the training set. We design an augmentation algorithm over the virtual sentences sampled from both of the vicinity distributions in sequence-to-sequence NMT model training. Experimental results on Chinese-English, English-French and English-German translation tasks demonstrate the capability of our approach to improving both translation performance and robustness.

\bibliographystyle{acl_natbib}
\balance
\bibliography{acl2020}
\end{document}